\documentclass[conference]{IEEEtran}
\IEEEoverridecommandlockouts
\usepackage{cite}
\usepackage{amsmath,amssymb,amsfonts}
\usepackage{algorithmic}
\usepackage{graphicx}
\usepackage{subcaption}
\usepackage{tabu}
\usepackage{textcomp}
\usepackage{xcolor}
\usepackage{multirow}
\newcommand{\etal}{\textit{et al}.}
\newcommand\T{\rule{0pt}{2.9ex}}       
\newcommand\B{\rule[-1.2ex]{0pt}{0pt}} 

\def\BibTeX{{\rm B\kern-.05em{\sc i\kern-.025em b}\kern-.08em
    T\kern-.1667em\lower.7ex\hbox{E}\kern-.125emX}}
\begin{document}

\title{Siamese Neural Networks for Skin Cancer Classification and New Class Detection using Clinical and Dermoscopic Image Datasets}

\author{\IEEEauthorblockN{Michael Luke Battle}
\IEEEauthorblockA{\textit{School of Computing} \\
\textit{Newcastle University, UK}\\
mail@lukebattle.com
}
\and
\IEEEauthorblockN{Amir Atapour-Abarghouei}
\IEEEauthorblockA{\textit{Department of Computer Science} \\
\textit{Durham University, UK}\\
amir.atapour-abarghouei@durham.ac.uk
}
\and
\IEEEauthorblockN{Andrew Stephen McGough}
\IEEEauthorblockA{\textit{School of Computing} \\
\textit{Newcastle University, UK}\\
stephen.mcgough@newcastle.ac.uk
}
}


\maketitle

\begin{abstract}
Skin cancer is the most common malignancy in the world. Automated skin cancer detection would significantly improve early detection rates and prevent deaths. To help with this aim, a number of datasets have been released which can be used to train Deep Learning systems -- these have produced impressive results for classification. However, this only works for the classes they are trained on whilst they are incapable of identifying skin lesions from previously unseen classes, making them unconducive for clinical use. We could look to massively increase the datasets by including all possible skin lesions, though this would always leave out some classes. Instead, we evaluate Siamese Neural Networks (SNNs), which not only allows us to classify images of skin lesions, but also allow us to identify those images which are different from the trained classes -- allowing us to determine that an image is not an example of our training classes. We evaluate SNNs on both dermoscopic and clinical images of skin lesions. We obtain top-1 classification accuracy levels of 74.33\% and 85.61\% on clinical and dermoscopic datasets, respectively. Although this is slightly lower than the state-of-the-art results, the SNN approach has the advantage that it can detect out-of-class examples. Our results highlight the potential of an SNN approach as well as pathways towards future clinical deployment.
\end{abstract}


\section{Introduction}

Not only is skin cancer the most common malignancy in the world, but its incidence rate is rising \cite{cakirEpidemiologyEconomicBurden2012,mishraOverviewMelanomaDetection2016}. Early detection can significantly improve the long term outcome, thus drastically reducing the mortality rate \cite{freemanAlgorithmBasedSmartphone2020}. Deep Learning (DL) based identification of skin cancer from images has shown considerable efficacy \cite{kassem2021machine, bevan2021skin, datta2021soft, liuDeepLearningSystem2020, chaturvediMulticlassSkinCancer2020, bevan2022detecting, jinnaiDevelopmentSkinCancer2020}. The best use of such DL-based systems would be by patients self-monitoring using an app such as \emph{MySkinSelfie}~\cite{hamptonUsabilityTestingMySkinSelfie2020}, where the system is able to provide a first level of triage. Therefore, there is a need for a robust DL solution where patients can photograph any skin lesions they may have -- many of which will be unrelated to cancer. Current solutions are not robust enough to be used in such an open environment. The common datasets used for training \cite{tschandlHAM10000DatasetLarge2018,
combaliaBCN20000DermoscopicLesions2019, codellaSkinLesionAnalysis2018, pachecoPADUFES20SkinLesion2020} contain samples for a small subset of skin lesion types, limiting all prognoses to these classes -- even for non-medical images. Likewise, it would be beneficial to be able to identify that a lesion wasn't one of these classes.

We aim to develop a deep learning system (DLS) that can automatically diagnose skin cancer using images taken with a smartphone camera. This model could be integrated into \emph{MySkinSelfie} \cite{hamptonUsabilityTestingMySkinSelfie2020}. Adding a reliable diagnostic feature to this app would enable patients to self-examine suspicious skin lesions. The prediction from the DLS could be shared with the patient and their dermatologist. This new process of skin cancer diagnosis would improve upon the current system in two ways. Firstly, a mobile application is more accessible than a general practitioner (GP), which would encourage more people to monitor their skin and improve early detection rates \cite{choudhuryEffectivenessSkinCancer2012}. Secondly, it would alleviate the burden of initial diagnoses from GPs and streamline the healthcare system \cite{udreaAccuracySmartphoneApplication2020}. 

Skin lesions can be categorised as cancerous or non-cancerous. Cancerous skin lesions fit into one of two groups, Malignant Melanoma (MEL) and Non-Melanoma Skin Cancer (NMSC). MEL is the most deadly form, accounting for approximately 75\% of all skin cancer-related deaths \cite{chaturvediMulticlassSkinCancer2020}. Due to its high mortality rate, previous work has approached skin cancer classification as a binary classification task, with images categorised as MEL or benign \cite{estevaDermatologistlevelClassificationSkin2017, brinkerDeepNeuralNetworks2019a, haenssleManMachineDiagnostic2018, nasr-esfahaniMelanomaDetectionAnalysis2016}. However, whilst MEL is far more deadly, the incidence rate of NMSC is significantly higher, accounting for 96\% of cases \cite{celebiEarlyDetectionMelanoma2015}. Moreover, different types of NMSC possess contrasting prognoses. For instance, the majority of cases of Basal Cell Carcinoma (BCC) are not life-threatening, but Squamous Cell Carcinoma (SCC) is responsible for 75\% of deaths within NMSC \cite{didonaNonMelanomaSkin2018}. It is therefore important to ensure that different types of NMSC are also diagnosed by the DLS. Additionally, it is required that the  DLS is able to distinguish previously unseen skin lesion types from the trained classes. Thus reducing the chance that non-cancerous lesions are misclassified as cancer. 

In constrained and controlled research environments, DL methods are effective at performing multi-class image classification of skin lesions \cite{harangiSkinLesionClassification2018, jinnaiDevelopmentSkinCancer2020, krohlingSmartphoneBasedApplication2021}. They have even outperformed dermatologists in their diagnostic accuracy \cite{chaturvediMulticlassSkinCancer2020}. However, currently available mobile applications have yet to translate this accuracy from a research environment into a practical setting \cite{ngooEfficacySmartphoneApplications2018, singhRecentAdvancementEarly2019}. A primary reason for this is that data used for training and testing is often composed of a relatively small number of classes, when compared to the number of potential types of skin lesions. For instance, a popular dataset for skin cancer classification is the HAM10000 dataset \cite{tschandlHAM10000DatasetLarge2018}. This is composed of seven skin lesion classes. However, Liu \etal \cite{liuDeepLearningSystem2020} noted 419 types of skin conditions. Therefore, HAM10000 is not reflective of the wide array of skin lesions that a mobile application would be exposed to in a practical setting, and to the best of our knowledge, there are no currently available datasets that are. This means that if a CNN exclusively trained on the HAM10000 dataset was used practically, previously unseen classes would be incorrectly diagnosed as one of the seven skin cancer classes. 

Open set recognition is a branch of DL research that investigates the possibility of detecting new unseen classes. Previous work has implemented techniques featuring OpenMax classifiers \cite{bendaleOpenSetDeep2015}, Generative Adversarial Nets \cite{nealOpenSetLearning2018} and Deep Open Classifiers \cite{shuDOCDeepOpen2017}. However, results in this area are far from optimal \cite{boultLearningUnknownSurveying2019}. We will take an alternative approach by using a Siamese Neural Network (SNN). A SNN uses neural networks to create an embedding, or a vectorised representation, of each image. During training, a triplet loss function is used to minimise the distance between embeddings of the same class, whilst pushing embeddings from different classes further apart \cite{liuSiameseConvolutionalNeural2019}. This creates clusters within the embedding space for each class, which can be used to classify new images using its $k$-nearest neighbours (KNN) \cite{pedregosaScikitlearnMachineLearning2011}. Likewise, using the position of unknown classes in the embedding space in relation to the clusters for trained classes, we are able to identify new classes -- as these do not lie close to the clusters for the known classes. As a proof of concept, we will first evaluate the capabilities of the models to classify images of faces as new. We will then seek to identify new skin lesion types. Before this can be performed, the classification performance of the framework must first be assessed. We perform this task on three datasets, two of which contain dermoscopic images and one with clinical images. Whilst the dataset with clinical images is a more realistic test for the mobile platform functionality, it has a limited number of images of skin cancers, in particular MEL. The dermoscopic datasets will thus provide a more rigorous test of the model's ability to classify skin cancer images.

\section{Related Work}
\label{main:background}

Deep learning systems previously used for skin cancer classification tasks mostly use deep CNNs which can only output classes that they have been trained on \cite{ chaturvediMulticlassSkinCancer2020,brinkerDeepNeuralNetworks2019, harangiSkinLesionClassification2018, liuDeepLearningSystem2020}. To address this issue, this paper explores an alternative approach based on the work of Schroff \etal \cite{schroffFaceNetUnifiedEmbedding2015} who pioneered the use of triplet loss in SNNs for face verification and achieved state-of-the-art levels of accuracy.

Ahmad \etal \cite{ahmadDiscriminativeFeatureLearning2020} applied this methodology for skin disease classification. They fine-tuned pre-trained ResNet-152 (RN) \cite{heDeepResidualLearning2015} and InceptionResNet-V2 (IRN) \cite{szegedyInceptionv4InceptionResNetImpact2016} models using the triplet loss function to create embeddings of skin disease images. They then computed the $L_2$ distances between these embeddings to classify the images. Whilst this paper applied this methodology to images of acne, spots, blackheads and dark circles, as opposed to forms of skin cancer, it still demonstrates the potential of this framework by achieving accuracy and sensitivity scores of 87.42\% and 97.04\%, respectively. Our work contrasts this prior work not only in the skin diseases classified, but also because they do not investigate if the network can effectively identify new classes.


There is limited research that explores the open set recognition aspect of SNNs with triplet loss. Previous work has used SNNs with a contrastive loss for anomaly detection \cite{castellaniRealWorldAnomalyDetection2021, raoTransferableNetworkSiamese2022, masanaMetricLearningNovelty2018}. However, Geng \etal \cite{gengRecentAdvancesOpen2021} note that anomaly detection differs from open-set recognition in that it only requires the identification of a few outliers - as opposed to the detection of unknown unknown classes as in our work. To the best of our knowledge, the work of Vetrova \etal \cite{vetrovaHiddenFeaturesExperiments2018} is the only one that explicitly investigates using SNNs with triplet loss for this purpose. In their work, they used one-class classification techniques with SNNs to classify species of moths and identify a new class. To create their novel class, they withheld one of the moth classes from model training and appended it to the test data. Other moth species were then treated as one class, and Support Vector Machines (SVMs) were used to classify images as novel or otherwise.

In contrast to this work, we aim to diagnose specific skin diseases. Therefore, an adapted version of this method will be used that preserves multi-class granularity for evaluation. Researchers have also used algorithms such as KNN for their test data classification, as opposed to SVM as used by Vetrova \etal \cite{vetrovaHiddenFeaturesExperiments2018}. Both Liu \etal \cite{liuDeepFewShotLearning2019} and Wang and Wang \cite{wangPlantLeavesClassification2019} implemented KNN algorithms for the inference phases of their research. More specifically, in each case they are applied to classify the test data using SNNs and the trained embeddings. It has been noted that this can improve the classification performance of SVMs by minimising the number of hyperparameters requiring optimisation \cite{liuDeepFewShotLearning2019}.

Previous work has approached skin cancer classification as a binary classification problem, where images are identified as melanoma or benign \cite{estevaDermatologistlevelClassificationSkin2017, brinkerDeepNeuralNetworks2019, nasr-esfahaniMelanomaDetectionAnalysis2016}. In this paper, we approach the problem as multi-class classification, and endeavour to diagnose images of skin cancer types. Krohling \etal \cite{krohlingSmartphoneBasedApplication2021} used a ResNet50 network to classify the PAD-UFES-20 dataset, achieving an impressive accuracy of 85\%. Harangi \cite{harangiSkinLesionClassification2018} implemented an ensemble of deep CNNs for three-class classification, obtaining an accuracy of 86.9\% on a dataset of dermoscopic images. Chaturvedi \etal \cite{chaturvediMulticlassSkinCancer2020} also classified dermoscopic images, performing seven-class classification on the HAM10000 dataset - which is also used here. They used deep CNNs such as Inception-V3 \cite{szegedyRethinkingInceptionArchitecture2015} and IRN \cite{szegedyInceptionv4InceptionResNetImpact2016}, comparing the results. The highest accuracy that they achieved was 93.2\%. The methods used in these papers differ from what is used in our work for the classification task, and whilst the results are impressive, these frameworks cannot handle images of a new class. This is one of the contributing factors as to why there has been limited success when attempting to replicate these results in clinical settings \cite{ngooEfficacySmartphoneApplications2018,singhRecentAdvancementEarly2019}. 

Current research in the area is directed at addressing this lack of applicability to the clinical setting. Groh \etal \cite{grohEvaluatingDeepNeural2021} demonstrated that darker skin colours are under-represented in most skin cancer datasets and verified that this would negatively impact model performance when classifying darker skin types. Therefore, any mobile application made available for public use must ensure it performs equally well with different skin tones. Moreover, \emph{Google} have recently developed their own mobile application, \emph{Google Health} \cite{liuDeepLearningSystem2020}. They applied deep learning systems to classify not just images of skin cancer, but 419 types of skin lesions. This research takes an alternative approach to our methodology, by using an Inception-V4 \cite{szegedyInceptionv4InceptionResNetImpact2016} model to provide a differential diagnosis across 27 of the most common diagnoses.

This is opposed to outputting a single diagnosis like the models developed by Krohling \etal \cite{krohlingSmartphoneBasedApplication2021}, Harangi \cite{harangiSkinLesionClassification2018} and Chaturvedi \etal \cite{chaturvediMulticlassSkinCancer2020}. It also provides a secondary diagnosis across 419 types of skin disease. Whilst this framework differs from our work, it overcomes the problem of new classes by training the model on significantly more classes than any previous work. The 27 classes that are used for the primary diagnosis contain 80\% of skin conditions seen in primary care. An additional discriminating feature between their work and ours is that their system uses a patient's medical history in the classification task. Their framework obtained top-1 accuracy and sensitivity scores of 71\% and 58\%, respectively. They also build on the work of Groh \etal \cite{grohEvaluatingDeepNeural2021} by ensuring a range of Fitzpatrick skin types are represented in both the training and validation data for the model, reporting top-1 accuracies of between 70\% and 74\% for types II-V \cite{fitzpatrickSoleilPeau1975}.

In this paper, we build upon the successes of past skin cancer classification systems and SNN to take an alternative approach towards skin cancer identification with open set recognition capabilities.

\section{Methodology}
\label{main:method}

This section describes the process for training the SNN, along with how the embedding space is used for both classification and identification of new classes.

\subsection{SNN Training}

\begin{figure*}[!t]
	\centering
	\includegraphics[scale=0.6]{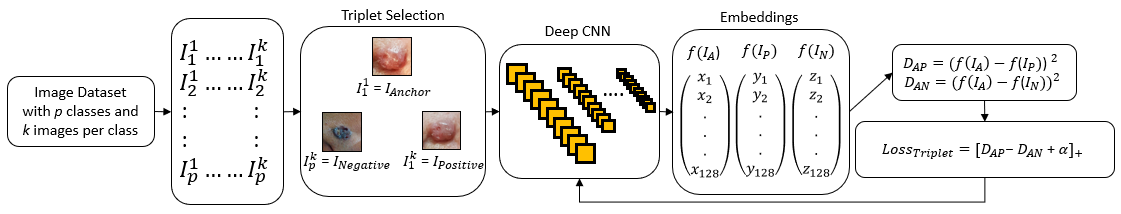}\hfill
	\caption{Method for SNN training based on \cite{ahmadDiscriminativeFeatureLearning2020}}
	\label{fig:network_architecture}
\end{figure*}

The SNN creates a 128-dimensional feature vector, or \emph{embedding}, for each image. This length has been shown to be optimal for classification accuracy \cite{schroffFaceNetUnifiedEmbedding2015}. A CNN is used for this task due to their invariance to geometric distortions and aptitude for identifying features in images \cite{lecunGradientbasedLearningApplied1998}. This CNN is then trained using triplet loss (as shown in Figure \ref{fig:network_architecture}), 
allowing us to adjust the sampling strategy and create more distinct clusters \cite{wuSamplingMattersDeep2018}. 
As the SNN is trained, embeddings for the same class move closer, whilst different classes are pushed apart. 

\subsubsection{Triplet Loss and Triplet Selection} To limit over-fitting, a network that minimises validation loss during training is used for final testing. We evaluate two online triplet mining (selection) methods: batch-all and batch-hard \cite{hermansDefenseTripletLoss2017}. Batch-hard uses only the hardest positive and negative for each
sample in a batch, whereas batch-all computes every possible triplet in a batch.  Each is assessed for classification accuracy and clustering ability. The latter will be assessed by decomposing the trained embedding space to PCA plots. 

\subsubsection{Datasets and Preprocessing} 
We use three publicly available datasets; the ISIC2019, HAM10000 and PAD-UFES-20. Each dataset contains cancerous and non-cancerous skin lesion images. A subset of 530 images from Labelled Faces in the Wild (LFW) \cite{huangLabeledFacesWild2007} is used to represent new classes. All data is anonymised, removing ethical concerns. ISIC2019 contains 25,998 dermoscopic images from the HAM10000 \cite{tschandlHAM10000DatasetLarge2018}, BCN20000 \cite{combaliaBCN20000DermoscopicLesions2019} and MSK datasets \cite{codellaSkinLesionAnalysis2018}. As different image preparation methods were used for each dataset, the images in ISIC2019 vary in both resolution and noise \cite{gessertSkinLesionClassification2020}. Due to its uniformity in comparison to ISIC2019, we also use the HAM10000 individually. This contains 10,015 dermoscopic images. The PAD-UFES-20 dataset \cite{pachecoPADUFES20SkinLesion2020} contains 2,298 clinical images. PAD-UFES-20 also provides information on the Fitzpatrick skin types for images in the dataset. 

A similar pre-processing method is applied for each dataset. Following the approach taken by Vetrova \etal \cite{vetrovaHiddenFeaturesExperiments2018}, we remove a class from the training data and append it to test data to function as a new class. We create the Test A and Test B datasets in this way, where Test A does not include the new class and Test B does. In PAD-UFES-20, we use Sebhorreic Keratoses (SEK) as the new class. Melanoctyic Nevi contextualises MEL during model training and Acitinic Keratoses can be a precursor to SCC so should be diagnosed \cite{tschandlHAM10000DatasetLarge2018}. Therefore, SEK is the most suitable lesion to function as the new class. For preparation of the ISIC2019 and HAM10000 datasets, we use non-cancerous Vascular Skin Lesions (VASC) as the novel class. 
Test C is a combination of Test A plus the 530 images from LFW. These images operate as the novel class in a proof-of-concept example as these images differ from skin lesions, but are not radically different. 

An additional step taken for the PAD-UFES-20 dataset is the removal of the fourth transparency channel from the images. Each image is then resized to $128\times 128$ pixels. 
This image resolution provides good performance for skin lesion classification, whilst lower resolutions impede model performance \cite{mahbodInvestigatingExploitingImage2021}. Each dataset is then normalised and split, where 80\% is used for model training and 20\% for evaluation. To alleviate the class imbalance problem, we augment the classes in the training data with horizontal and vertical flips, as performed in previous research \cite{harangiSkinLesionClassification2018, jinnaiDevelopmentSkinCancer2020,  ahmadDiscriminativeFeatureLearning2020}. Class weighting and downsampling are also tested as possible solutions.

\subsubsection{Models for Embedding Layer} We test several CNNs as our embedding layer: InceptionResNet-V2 (IRN) \cite{szegedyInceptionv4InceptionResNetImpact2016}, ResNet-152 (RN) \cite{heDeepResidualLearning2015} and EmbeddingNet (EN) \cite{bielskiSiameseTripletLearning2021}. IRN and RN produced good results when employed by Ahmad \etal \cite{ahmadDiscriminativeFeatureLearning2020} for classifying skin lesions using SNNs. RN utilises shortcut connections from ResNet \cite{heDeepResidualLearning2015} but in a far deeper network. IRN combines these shortcut connections with the Inception blocks first used by He \etal \cite{heDeepResidualLearning2015,szegedyGoingDeeperConvolutions2014}. EN contains only two convolutional layers. As the network is intended to function on mobile devices, EN will assess the performance trade-off of a computationally lighter model for the classification task. 

To optimise the CNNs, we vary individual parameters and observe their results. We use transfer learning \cite{bozinovskiReminderFirstPaper2020} with IRN and RN, where the models are pre-trained on the ImageNet dataset \cite{russakovskyImageNetLargeScale2015}. All weights in the network are then fine-tuned as recommended in previous work \cite{ahmadDiscriminativeFeatureLearning2020, anwarMedicalImageAnalysis2018}. However, Hermans \etal \cite{hermansDefenseTripletLoss2017} note that pre-trained embedding layers can reduce model flexibility. Therefore, we compare the classification performance of this approach to when the weights are randomly initialised. The effect of $L_2$ normalisation in the final layer on classification performance is also examined \cite{schroffFaceNetUnifiedEmbedding2015}. 

We implement stochastic gradient descent \cite{wilsonGeneralInefficiencyBatch2003} with a momentum of 0.8, as used by Ahmad \etal \cite{ahmadDiscriminativeFeatureLearning2020}. Adam \cite{kingmaAdamMethodStochastic2017} is also trialled with default parameters, as recommended by Hermans \etal \cite{hermansDefenseTripletLoss2017}. Finally, the model is trained using a batch size of 128. This ensures there are valid triplets for each class in each batch, while larger batches exceed hardware limitations.

\subsection{Classifying Dermoscopic and Clinical Images}

Before evaluating the model's ability to discern new classes, it is important to first evaluate its classification performance. 
Once the model has been trained, we generate embeddings of the images from Test A. 
The KNN algorithm is then used to assign them a class based on which training embeddings they are closest to in the 128-dimensional hyperspace. To ensure that our results are optimal, we evaluate $k \in \{1,2,..,50\}$. The value of $k$ that delivers the best classification accuracy will be used in the final model. 

\subsection{Identification of New Classes}
\label{main:new_classes}

\begin{figure}[!t]
\centering
\begin{subfigure}{.25\textwidth}
  \centering
  \includegraphics[width=\textwidth]{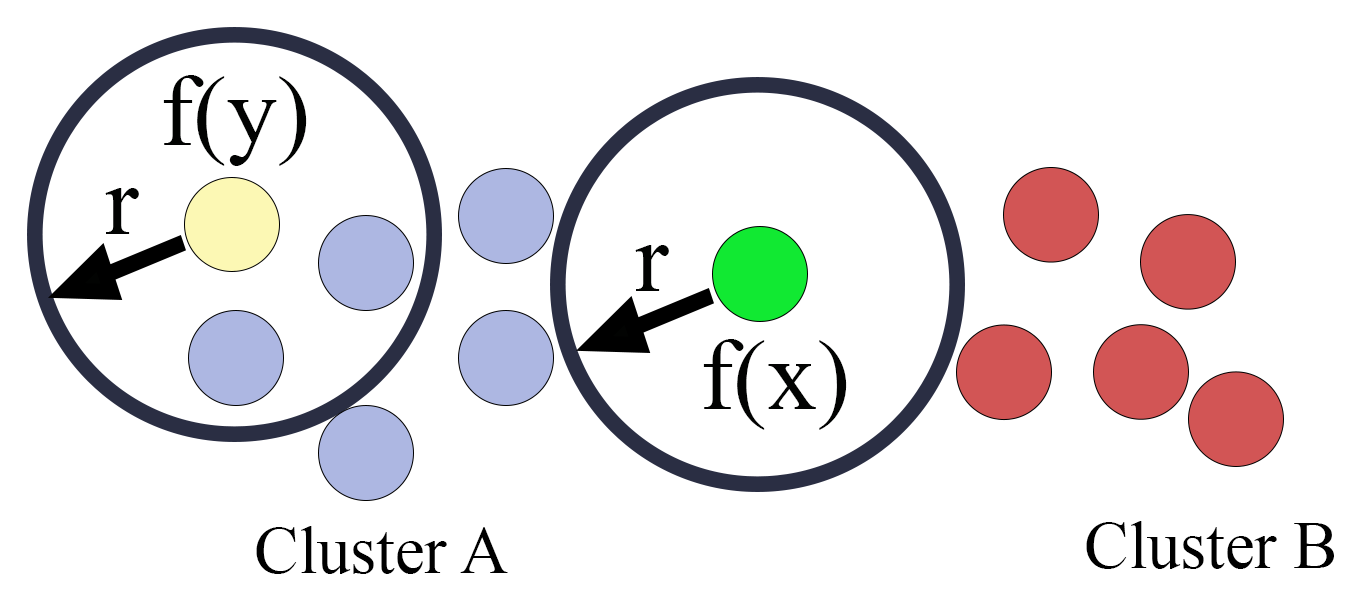}  
  \caption{Distance-based method}
  \label{fig:distance-1}
\end{subfigure}%
\begin{subfigure}{.25\textwidth}
  \centering
  \includegraphics[width=\textwidth]{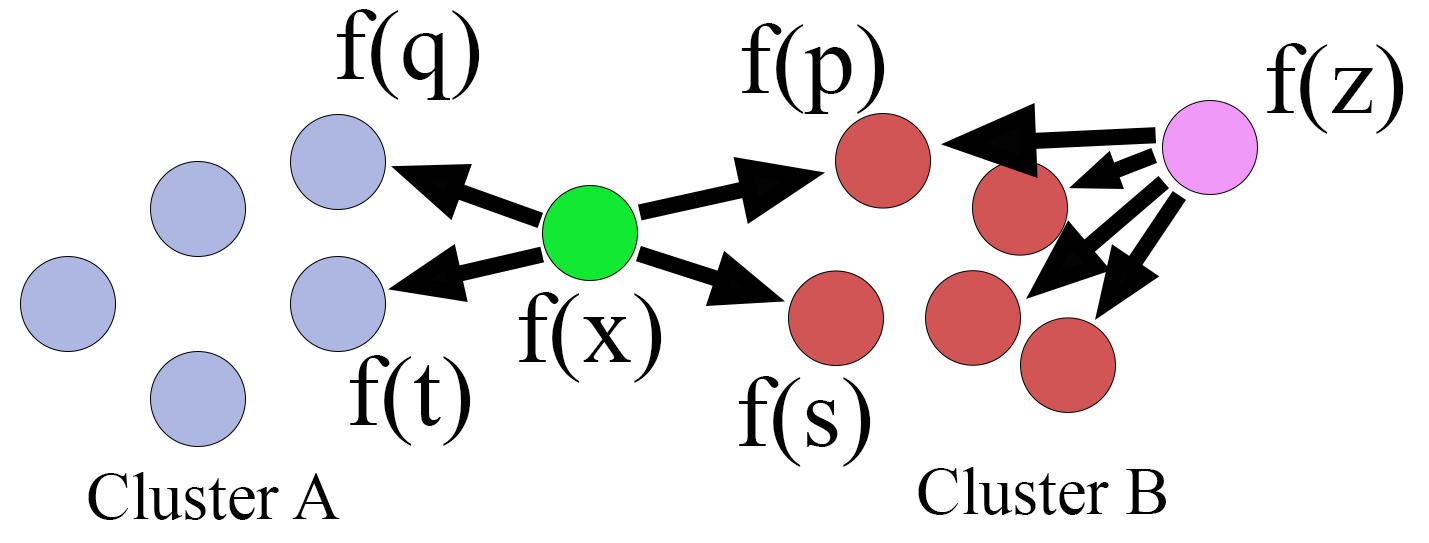}  
  \caption{Probability-based method}
  \label{fig:certainty-1}
\end{subfigure}
\caption{$f(x), f(y), f(z), f(q), f(p), f(t)$ and $f(s)$ are $n$-dimensional embeddings $f(x),f(y),f(z), f(q)$, $f(p), f(t), f(s) \in \mathbb{R}^n$ generated from the transformative embedding layer $f$, such that $n \in \mathbb{N}$. $x$, $y$ and $z$ are test images. Clusters A and B represent $n$-dimensional training embeddings generated by $f$ of classes A and B, respectively. $r$ represents a given radius such that $r \in \mathbb{R}_{+} $. Test images $y$ and $z$ are classified as classes A and B, respectively. In Figure \ref{fig:distance-1}, $x$ is classified as new due to no training embeddings existing within radius $r$ of $f(x)$. In Figure \ref{fig:certainty-1}, $x$ is classified as new if the probability threshold is 0.6. }\vspace{-0.2cm}
\label{}
\end{figure}


To the best of our knowledge, this paper is the first to rigorously evaluate this property of SNNs for multi-class classification. We implement two methods to identify new classes, using distance and probability. The method using distance is shown in Figure \ref{fig:distance-1}. Each test embedding is classified by majority vote using the training embeddings that fall within a radius $r$. If no embeddings fall within this distance, as with $f(x)$, this is identified as a new class. Similarly, $f(y)$, would be classified as part of cluster A. 

When using probability, novel classes are identified by the number of different classes closest to them. KNN finds the closest $k$ training embeddings for each test embedding and generates class probabilities using the proportion of each class in the $k$ embeddings. Note that other algorithms such as mean shift \cite{yizongchengMeanShiftMode1995} or DBSCAN \cite{esterDensityBasedAlgorithmDiscovering1996} could be used instead of KNN.  
In Figure \ref{fig:certainty-1}, when $k = 4$, $P(f(x)\in A)=P(f(x)\in B)=0.5$ where $P(f(x) \in A)$ represents the probability that $f(x)$ is in class $A$. Therefore, in this example, if we set a maximum probability threshold to $P_{threshold}=0.6$, then $P(f(x) \in A), P(f(x) \in B) < P_{threshold}$ so $f(x)$ would be identified as a new class. Similarly, embedding $f(z)$ would be classified as belonging to class B as $P(f(x) \in B)> P_{threshold}$.

\section{Results \& Evaluation}
\label{main:results}

\subsection{Evaluation Methods}

We first evaluate the model's classification performance using a variety of metrics with the Test A dataset. KNN is used to derive the top-1 and top-3 accuracy scores. These can be used in the mobile app to provide the three most likely diagnoses rather than just one, as implemented by Liu \etal \cite{liuDeepLearningSystem2020}. 
We also examine the sensitivity and specificity for each class. Sensitivity is of particular interest as it describes the models efficacy at detecting a positive diagnosis. 
This is particularly important for a model used by members of the public, as it could provide false reassurance to a patient with skin cancer. We will pay the most attention to the sensitivity rates of the most serious forms of skin cancer, MEL and SCC. In contrast, specificity identifies how frequently a patient is misdiagnosed as having a disease. This is also monitored as a false diagnosis may cause patients undue anxiety \cite{harangiSkinLesionClassification2018}.

The open set recognition capabilities of the proposed approach are also rigorously tested. The models and resulting embedding spaces with the best top-1 classification accuracy for each dataset are used to identify new classes. We use the Test C data to evaluate how well the embedding space can detect images of faces from LFW, and Test B for images of a new skin lesion class. We assess both methods of new class identification using the same process. Due to the class imbalance in each dataset and the variation in size of the new class, we monitor the sensitivity per class for $r, P_{threshold} \in (0,1]$ (Figure \ref{fig:pad_ufes_plots}). An optimal value is then selected to compute top-1 accuracy and the confusion matrix (Figure \ref{fig:pads_ufes_cf}). This value should prioritise skin cancer sensitivity to limit the number of images misclassified as new. The confusion matrix provides the count and percentage of class population. 
Using this confusion matrix, we derive two key metrics; the sensitivity of the new class and the average percentage of skin lesion classes misclassified as new (denoted as the MN score). The latter is calculated by taking the mean average of the percentages of each skin disease type classified as new. This facilitates the assessment of how well new classes are identified and the extent to which this has impeded classification performance. 




\subsection{Optimisation Results}

The final models use a learning rate of 0.0001, the Adam optimiser and $L_2$ normalisation in the final layer. These were found to be optimal in terms of top-1 accuracy and convergence speed. 
Transfer learning was shown to be effective, so we will use IRN and RN models pre-trained on ImageNet and tune all of the weights. Class weighting will not be used as this was shown to negatively impact model performance. Instead, data augmentation is applied to the under-represented classes and all images are used rather than downsampled, as this is shown to significantly improve performance.

\begin{figure}
\centering
\begin{subfigure}{.24\textwidth}
  \centering
  \includegraphics[width=\textwidth]{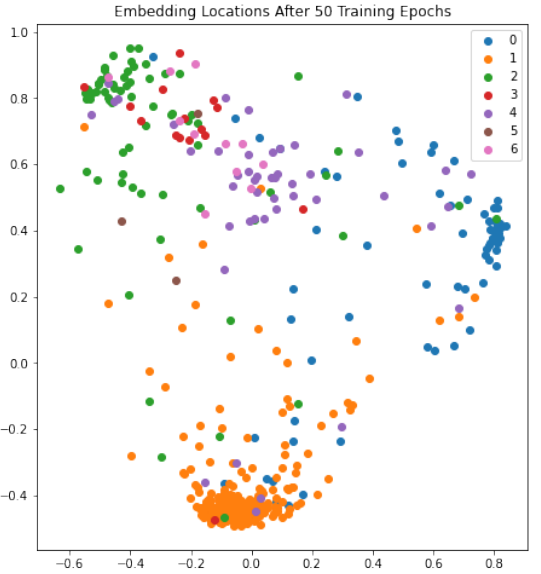}
  \caption{Batch-all}
  \label{fig:pca-batchall}
\end{subfigure}%
\begin{subfigure}{.26\textwidth}
  \centering
  \includegraphics[width=\textwidth]{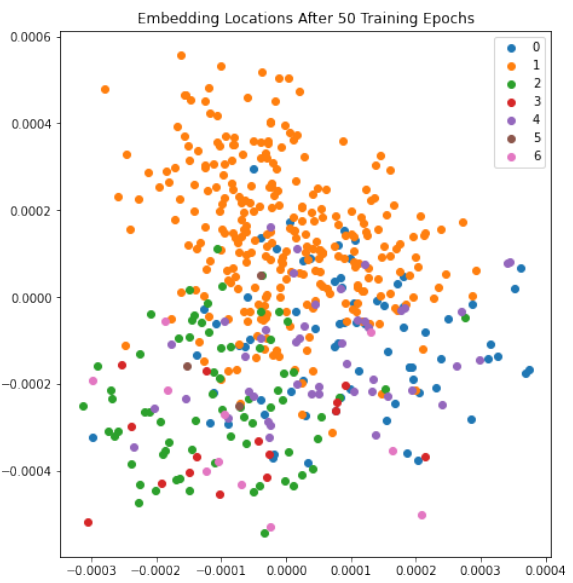}
  \caption{Batch-hard}
  \label{fig:pca-batchhard}
\end{subfigure}
\caption{Test embeddings with IRN trained using different triplet mining methods on the ISIC2019 dataset.}
\label{fig:pca}\vspace{-0.45cm}
\end{figure}

We also tested two online triplet mining strategies, batch-all and batch-hard, each with a margin of 0.2 in keeping with the literature recommendations for triplet loss \cite{schroffFaceNetUnifiedEmbedding2015, ahmadDiscriminativeFeatureLearning2020}. It was found that batch-all significantly outperforms batch-hard in terms of top-1 and top-3 classification accuracy. The PCA plots in Figure \ref{fig:pca} show that batch-all is more effective at creating distinct clusters, which is important for identifying new classes. One reason for this difference may be that during batch-hard triplet selection, outliers are selected as these would constitute the hardest triplets, causing the embeddings to converge to zero. Finally, a value of $k=26$ for KNN was found to be optimal for top-1 and top-3 accuracy, so this will be used to compute metrics for the remainder of this paper.

\subsection{Classification}

\subsubsection{Performance on ISIC2019}

\begin{table*}[!t]
\captionsetup[table]{skip=4pt}
\captionof{table}{Top-1 and top-3 accuracy (\%) for each model on the ISIC2019, HAM10000 and PAD-UFES-20 datasets. The highest metric for each dataset is emboldened.}
\label{table:acc-results}
	\centering
		{\tabulinesep=0mm
			\begin{tabu}{@{\extracolsep{3pt}}c c c c c c c@{}}
				\hline\hline
				\multicolumn{1}{c}{\multirow{2}{*}{Dataset}} & 
				\multicolumn{2}{c}{IRN} &
				\multicolumn{2}{c}{RN} &
				\multicolumn{2}{c}{EN}\T\B \\
				\cline{2-3} \cline{4-5} \cline{6-7}
						& Top-1 (\%) & Top-3 (\%) & Top-1 (\%) & Top-3 (\%) & Top-1 (\%) & Top-3 (\%)\T\B\\
				\hline\hline
ISIC2019   & \textbf{78.39}  & 87.05 & 77.19 & \textbf{88.26} & 64.49 & 87.92\T\\
HAM10000 & \textbf{85.16} & 94.03 & 82.23 & \textbf{94.84} & 73.06 & 93.11 \\
PAD-UFES-20 & 70.70 & 84.26 & \textbf{74.33} & \textbf{85.96} & 50.61 & 72.64 \B\\
\hline
\hline
\end{tabu}
}
\end{table*}

The classification results of EN, IRN and RN on the ISIC2019 dataset are shown in Tables \ref{table:acc-results} \& \ref{table:isic2019-results-2}. Table \ref{table:acc-results} shows 
that IRN has the highest top-1 accuracy with 78.39\%, whilst RN has the highest top-3 accuracy with 88.26\%. Table \ref{table:isic2019-results-2} 
shows that IRN outperforms RN for the non-cancerous classes and SCC, with RN performing better on MEL and BCC. In both cases, the sensitivity score for MEL is low at 62.50\% and 63.86\% for IRN and RN, respectively. However, these results are comparable to the sensitivity score of 59.40\% for MEL obtained by the winners of the ISIC2019 challenge \cite{gessertSkinLesionClassification2020}. It is also similar to the top-1 sensitivity score of 57\% for cancerous skin lesions obtained by Liu \etal \cite{liuDeepLearningSystem2020}.

An additional result to note from Table \ref{table:acc-results} is that whilst EN performs significantly worse than RN and IRN in terms of top-1 accuracy, it has a good top-3 result of 87.92\%. This is higher than the corresponding result with IRN of 87.05\% and only slightly lower than the result obtained by RN.

\begin{table*}[!t]
\captionsetup[table]{skip=4pt}
\captionof{table}{Sensitivity and Specificity of each model and class in the ISIC2019 dataset. The best sensitivity result per skin lesion type is emboldened. Note that the ISIC2019 winners used the ISIC2019 test dataset as well as one additional class (VASC) for their classification - so this comparison has limited meaning.}
\label{table:isic2019-results-2}
	\centering
		{\tabulinesep=0mm
			\begin{tabu}{@{\extracolsep{3pt}}c c c c c c c c c@{}}
				\hline\hline
				\multicolumn{1}{c}{\multirow{2}{*}{Skin Disease}} & 
				\multicolumn{2}{c}{IRN} &
				\multicolumn{2}{c}{RN} &
				\multicolumn{2}{c}{EN} &
				\multicolumn{2}{c}{ISIC2019 Winners \cite{gessertSkinLesionClassification2020}}\T\B \\
				\cline{2-3} \cline{4-5} \cline{6-7} \cline{8-9}
						& SE (\%) & SP (\%) & SE (\%) & SP (\%) & SE (\%) & SP (\%) & SE (\%) & SP (\%)\T\B\\
				\hline\hline
Malignant Melanoma (MEL)   & 62.50  & 95.70 & \textbf{63.86} & 93.93& 40.68 & 92.24 & 59.4 & 96.2\T\\
Melanocytic Nevi (NEV) & \textbf{91.31} & 85.35 & 89.38 & 85.51 & 89.15 & 70.82 & 71 & 97.5\\
Basal Cell Carcinoma (BCC) & 74.89 & 96.70 & \textbf{75.48} & 96.22 & 54.06 & 93.62 & 72.1 & 94 \\
Acitinic Keratoses (AK) & \textbf{58.70} & 97.54 & 54.35 & 97.95 & 40.76 & 95.72 & 48.4 & 96.5\\
Dermatofibroma (DF) & \textbf{63.27} & 99.23 & 51.02 & 99.74 & 0.00 & 99.98 & 57.8 & 99.2 \\
Benign Keratosis DF (BKL) & \textbf{57.40} & 96.52 & 57.20 & 96.04 & 19.80 & 96.52 & 39.4 & 98.5\\
Squamous Cell Carcinoma (SCC) & \textbf{58.73} & 98.04 & 49.21 & 98.16 & 19.05 & 97.79 & 43.9 & 98.6\B\\
\hline
\hline
\end{tabu}
}
\vspace{-5pt}
\end{table*}

\subsubsection{Performance on HAM10000}

Tables \ref{table:acc-results} \& \ref{table:ham10000-results-2} display the results of the six-class classification performed on the HAM10000 dataset. As with the results for the ISIC2019 dataset, IRN obtains the highest top-1 accuracy result with 85.16\%. RN also achieved the best top-3 accuracy score with 94.84\%. Comparatively, Chaturvedi \etal\  achieved a maximum of 93.20\% accuracy when performing classification on the full HAM10000 dataset \cite{chaturvediMulticlassSkinCancer2020}. Therefore, in terms of accuracy our approach is competitive. These results are also notably higher than the maximum top-1 accuracy obtained with the ISIC2019 dataset of 78.39\%. As noted in Section \ref{main:method}, the images in HAM10000 are far more uniform than those in ISIC2019 \cite{gessertSkinLesionClassification2020}. This consistency may be a contributing factor to the observed increase in accuracy, and an element worthy of consideration when creating the mobile application.

\begin{table*}[!t]
\captionsetup[table]{skip=4pt}
\captionof{table}{Sensitivity and Specificity of each model/class in HAM10000. Best sensitivity per lesion type is emboldened.}
\label{table:ham10000-results-2}
	\centering
		{\tabulinesep=0mm
			\begin{tabu}{@{\extracolsep{3pt}}c c c c c c c@{}}
				\hline\hline
				\multicolumn{1}{c}{\multirow{2}{*}{Skin Disease}} & 
				\multicolumn{2}{c}{IRN} &
				\multicolumn{2}{c}{RN} &
				\multicolumn{2}{c}{EN}\T\B \\
				\cline{2-3} \cline{4-5} \cline{6-7}
						& SE (\%) & SP (\%) & SE (\%) & SP (\%) & SE (\%) & SP (\%)\T\B\\
				\hline\hline
Malignant Melanoma (MEL)   & \textbf{59.82}  & 96.69 & 52.23 & 96.00 & 16.96 & 98.57 \T\\
Melanocytic Nevi (NEV) & \textbf{95.74} & 81.16 & 94.92 & 81.00 & 94.99 & 54.79 \\
Basal Cell Carcinoma (BCC) & \textbf{79.79} & 98.83 & 69.15 & 98.03 & 51.06 & 95.22 \\
Acitinic Keratoses (AK) & \textbf{76.12} & 98.01 & 52.24 & 97.64 & 28.36 & 97.48 \\
Dermatofibroma (DF) & 36.00 & 99.90 & \textbf{48.00} & 99.44 & 0.00 & 100.00 \\
Benign Keratosis DF (BKL) & \textbf{58.15} & 96.97 & 55.07 & 96.17 & 29.52 & 95.37 \B\\
\hline
\hline
\end{tabu}
}
\vspace{-0.35cm}
\end{table*}

However, Table \ref{table:ham10000-results-2} shows that despite this, the highest sensitivity achieved with MEL was 59.82\% with IRN. This is lower than the sensitivity result of 62.50\% obtained with this model with the ISIC2019 dataset. This may be due to 3,642 MEL images being used for model training when classifying the ISIC2019 dataset, and only 889 used with the HAM10000. This difference in training images for MEL may contribute to the decrease in sensitivity between the models. In terms of model performance, IRN has the highest sensitivity score for each skin lesion type, except Dermatofibroma. 

As with its result with the ISIC2019 dataset, EN performs well with HAM10000 in terms of top-3 accuracy with a score of 93.11\%. This result, along with the top-3 accuracy score on the ISIC2019 dataset, suggests that with further work this model could provide a computationally inexpensive option for performing top-3 classification of skin lesion datasets.

\subsubsection{Performance on PAD-UFES-20}

\begin{table*}[!t]
\captionsetup[table]{skip=4pt}
\captionof{table}{Sensitivity and Specificity of each model and class in the PAD-UFES-20 dataset. The best sensitivity result per skin lesion type is emboldened.}
\label{table:pads-results-2}
	\centering
		{\tabulinesep=0mm
			\begin{tabu}{@{\extracolsep{3pt}}c c c c c c c@{}}
				\hline\hline
				\multicolumn{1}{c}{\multirow{2}{*}{Skin Disease}} & 
				\multicolumn{2}{c}{IRN} &
				\multicolumn{2}{c}{RN} &
				\multicolumn{2}{c}{EN}\T\B \\
				\cline{2-3} \cline{4-5} \cline{6-7}
						& SE (\%) & SP (\%) & SE (\%) & SP (\%) & SE (\%) & SP (\%)\T\B\\
				\hline\hline
Malignant Melanoma (MEL)   & 41.67 & 99.00 & \textbf{66.67} & 99.00 & 0.00 & 100.00  \T\\
Melanocytic Nevi (NEV) & 75.56 & 97.83 & \textbf{88.89} & 96.20 & 71.11 & 90.22 \\
Basal Cell Carcinoma (BCC) & 72.19 & 74.18 & \textbf{80.47} & 81.56 & 56.80 & 72.54 \\
Acitinic Keratoses (AK) & \textbf{78.95} & 82.38 & 75.67 & 89.66 & 48.68 & 79.69 \\
Squamous Cell Carcinoma (SCC) & \textbf{31.43} & 97.62 & 22.86 &95.77  & 20.00 & 87.30 \B\\
\hline
\hline
\end{tabu}
}
\vspace{-5pt}
\end{table*}

The results of the five-class classification task performed on the PAD-UFES-20 clinical image dataset can be viewed in Tables \ref{table:acc-results} \& \ref{table:pads-results-2}. Table \ref{table:acc-results} shows that RN performs best in terms of both top-1 and top-3 accuracy, with scores of 74.33\% and 85.96\%, respectively. This top-1 accuracy is substantially higher than the IRN model score of 70.70\%, suggesting that depth is more important than width when our framework is applied to noisier images. 

However, this score is far lower than the accuracy score of 85\% obtained by Krohling \etal \cite{krohlingSmartphoneBasedApplication2021} when classifying PAD-UFES-20 using ResNet-50. They also utitlised patient information in their research, which may have contributed to the improved classification performance. Our results are also lower than the accuracies achieved by models operating on the ISIC2019 and HAM10000 datasets in the previous sections. This is expected due to the increased noise present in clinical images when compared to dermoscopic images. Although there were fewer classes to classify in PAD-UFES-20 than the other datasets, the accuracy is still not far off, suggesting that this technique holds value.

Table \ref{table:pads-results-2} shows that RN has the highest sensitivity scores for the majority of lesion types, including MEL. It also highlights that EN performs poorly with a sensitivity score of 0\% for MEL. Table \ref{table:acc-results} also shows that EN obtained top-1 and top-3 accuracy scores of 50.61\% and 72.64\% on PAD-UFES-20, respectively. This is far lower than results on HAM10000 and ISIC2019, indicating that EN could be implemented as a top-3 classifier for dermoscopic but not clinical images.

\subsection{Performance with New Classes in PAD-UFES-20}
\label{main:new_classe_results}


\subsubsection{Identifying Faces From LFW}

\begin{figure*}[!t]
\centering
\begin{subfigure}{.4\textwidth}
  \centering
  \includegraphics[width=.8\linewidth]{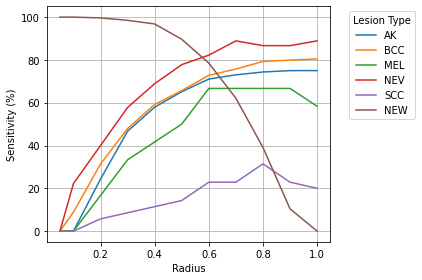}  
  \caption{DT-LFW}
  \label{fig:distance_metric_pads_ufes_lfw}
\end{subfigure}
\begin{subfigure}{.4\textwidth}
  \centering
  \includegraphics[width=.8\linewidth]{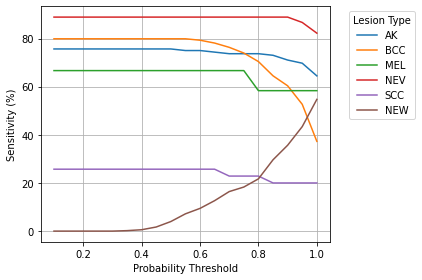}  
  \caption{PT-LFW}
  \label{fig:prob_metric_pads_ufes_lfw}
\end{subfigure}
\begin{subfigure}{.4\textwidth}
  \centering
  \includegraphics[width=.8\linewidth]{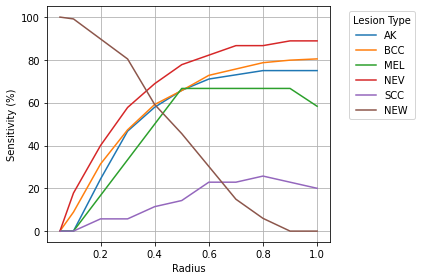}  
  \caption{DT-SEK}
  \label{fig:distance_metric_pads_ufes}
\end{subfigure}
\begin{subfigure}{.4\textwidth}
  \centering
  \includegraphics[width=.8\linewidth]{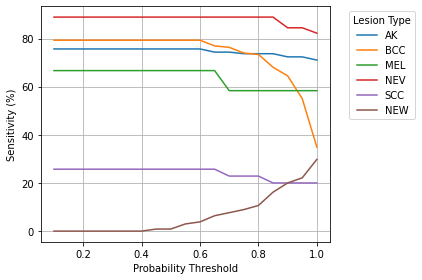}  
  \caption{PT-SEK}
  \label{fig:prob_metric_pads_ufes}
\end{subfigure}
\caption{Sensitivity per type. DT and PT are distance and probability thresholds. LFW / SEK denote where new class is from.}
\label{fig:pad_ufes_plots}\vspace{-0.35cm}
\end{figure*}

\begin{figure*}[!t]
\centering
\begin{subfigure}{.4\textwidth}
  \centering
  \includegraphics[width=.8\linewidth]{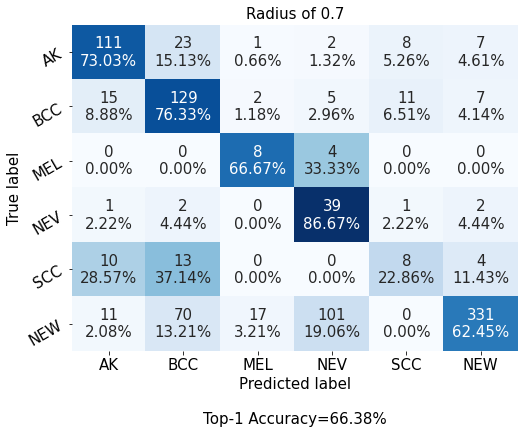}  
  \caption{DT-LFW}
  \label{fig:distance_metric_pads_ufes_lfw_cf}
\end{subfigure}
\begin{subfigure}{.4\textwidth}
  \centering
  \includegraphics[width=.8\linewidth]{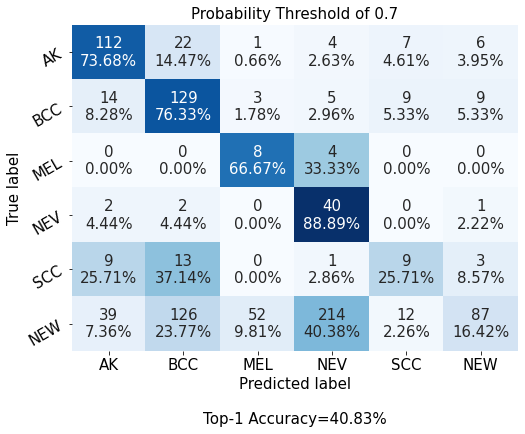}  
  \caption{PT-LFW}
  \label{fig:prob_metric_pads_ufes_lfw_cf}
\end{subfigure}
\begin{subfigure}{.4\textwidth}
  \centering
  \includegraphics[width=.8\linewidth]{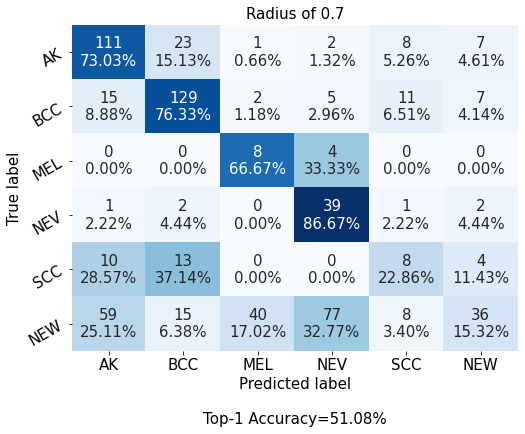}  
  \caption{DT-SEK}
  \label{fig:distance_metric_pads_ufes_cf}
\end{subfigure}
\begin{subfigure}{.4\textwidth}
  \centering
  \includegraphics[width=.8\linewidth]{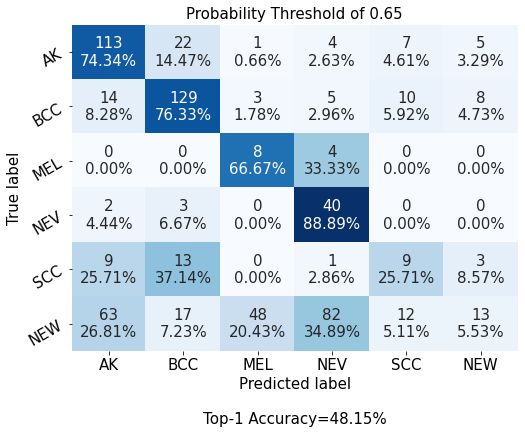}  
  \caption{PT-SEK}
  \label{fig:prob_metric_pads_ufes_cf}
\end{subfigure}
\caption{Confusion matrices where DT and PT denote distance and probability thresholds, respectively. LFW and SEK mean that the new class is taken from LFW or is an image of SEK, respectively.}
\label{fig:pads_ufes_cf}
\end{figure*}

Before evaluating if new lesion types can be detected, we test if the trained embedding space can identify faces from LFW as new classes. Figure \ref{fig:distance_metric_pads_ufes_lfw} shows that the sensitivity of the new class decreases slowly as the radius is increased. We observe that at a radius of 0.7, the sensitivities of the skin lesion classes are close to their maximum values. Figure \ref{fig:distance_metric_pads_ufes_lfw_cf} indicates that 62.45\% of the new classes are correctly identified. This suggests that the embeddings of the new class are mostly distinct from the skin lesion classes in hyperspace. Additionally, we derive an MN score of 4.92\%. These results highlight the potential for this framework, as a majority of new classes have been successfully identified without hindering classification performance.

Varying the probability threshold produces different results. Figure \ref{fig:prob_metric_pads_ufes_lfw} shows that the sensitivity of the new class increases with the probability threshold, while the other class sensitivities decrease. Skin lesion class sensitivities are still high at 0.7. Figure \ref{fig:prob_metric_pads_ufes_lfw_cf} shows the confusion matrix at this value and we observe a low MN score of 4.01\%. However, only 16.42\% of new class images have been identified - significantly less than the 62.45\% detected with a distance threshold. This shows that for these new class images, the distance threshold is far more effective at identifying new classes. 




\subsubsection{Using Distance to Identify New Skin Lesion Classes}


Figure \ref{fig:distance_metric_pads_ufes} shows that the sensitivity of the new class, consisting of images of the previously unseen class - SEK, falls significantly faster as the radius is increased when compared to Figure \ref{fig:distance_metric_pads_ufes_lfw}. This indicates that the SEK embeddings are closer to the training embeddings in hyperspace than the images from LFW. This is further supported by new class sensitivity and MN scores in Figure \ref{fig:distance_metric_pads_ufes_cf} of 15.32\% and 4.92\%, respectively. Clearly, this is far less than the new class sensitivity of 62.45\% obtained when implementing a distance threshold to identify images from LFW. These results show that the model can distinguish new classes that are sufficiently distinct, but more work must be done to ensure the model can also effectively identify previously unseen skin lesion types.


\subsubsection{Using Probability to Identify New Skin Lesion Classes}


A probability threshold performs worse than a distance threshold when identifying previously unseen images of SEK. Figure \ref{fig:prob_metric_pads_ufes_cf} highlights that only 5.53\% are correctly identified, far less than the score of 15.32\% achieved with a distance threshold and the same data. This, combined with the outcome of the previous section, indicates that the distance threshold is the better technique for identifying new classes in this case. This result is also less than the new class sensitivity of 16.42\% achieved by the same method with the LFW images in the previous section, reiterating that new skin lesion embeddings are less distinct in hyperspace than the images from LFW. 

\subsubsection{Comparison of Datasets}

\begin{table*}[!t]
\captionsetup[table]{skip=4pt}
\captionof{table}{The new class sensitivity (NCS) and average percentage misclassified as new (MN), for both methods in datasets.}
\label{table:new_class_summary}
	\centering
		{\tabulinesep=0mm
			\begin{tabu}{@{\extracolsep{3pt}}c c c c c c@{}}
				\hline\hline
				\multicolumn{1}{c}{\multirow{2}{*}{Method}} &
				\multicolumn{1}{c}{\multirow{2}{*}{Dataset}} &
				\multicolumn{2}{c}{LFW as New Class} &
				\multicolumn{2}{c}{Skin Lesion as New Class} \T\B \\
				\cline{3-4} \cline{5-6}
						& & NCS (\%) & MN (\%) & NCS (\%) & MN (\%)\T\B\\
				\hline\hline
\multirow{3}{6em}{Distance Threshold} & PAD-UFES-20  & 62.45  & 4.92 & 15.32 & 4.92 \T\\
& HAM10000  & 23.21 & 7.07 & 8.45 & 7.07  \\
& ISIC2019 & 2.08 & 5.61 & 11.86 & 5.61\B\\
\hline
\multirow{3}{6em}{Probability Threshold} & PAD-UFES-20  & 16.42  & 4.01 & 5.53 & 3.32\T\\
& HAM10000  & 10.00 & 10.94 & 8.45 & 10.94  \\
& ISIC2019 & 5.28 & 5.39 & 7.11 & 4.01 \B\\
\hline
\hline
\end{tabu}
}
\vspace{-10pt}
\end{table*}


It is clear from Table \ref{table:new_class_summary} that the best results for both types of new class were achieved using a distance threshold with the PAD-UFES-20 dataset. This result is interesting, as the highest classification accuracies were achieved by models classifying the HAM10000 and ISIC2019 datasets. However, the ISIC2019 and HAM10000 also have significantly more embeddings in the embedding space. This may be why new classes are identified less effectively in these datasets than in PAD-UFES-20, implying that sparsity is important when curating an embedding space for identifying new classes. Analysis of the HAM10000 and ISIC2019 datasets are not discussed further due to space limitations.

\subsection{Limitations}

There are several key limitations to the work produced in this paper. A significant portion of these are inherited from the datasets. For instance, in PAD-UFES-20 and HAM10000, only 58.4\% and 53.3\% of the skin lesion images have been biopsy-verified, respectively \cite{tschandlHAM10000DatasetLarge2018,pachecoPADUFES20SkinLesion2020}. Therefore, there is a chance that our models have inherited bias in the dermatologist-diagnosed images. Additionally, the HAM10000 and ISIC2019 datasets do not give a breakdown of the Fitzpatrick skin types and may be biased towards different skin tones. Finally, there is a significant class imbalance in each dataset that leads to varied model performance for different skin lesion types. This is a problem with real-world datasets for skin lesion classification that has been documented in prior work \cite{gessertSkinLesionClassification2020}.   

Another area where this work has limitations is in data preprocessing and model training. In each case the test data is a subset of the dataset, so before deploying a model, generalisation would require testing on truly external data \cite{brinkerDeepNeuralNetworks2019}. Cross-validation may have provided a more thorough test of this property. Additionally, data augmentation such as rotation and zooming could have been explored. Moreover, as our data preprocessing involves removing a non-cancerous class to function as the new class, the classification performance of each model has not been tested on the full datasets.

\section{Conclusions and Future Work}
\label{main:conclusion}

In this paper, we have trained several SNNs  using the triplet loss function and evaluated their ability to classify skin disease datasets and identify images of a new class. To the best of our knowledge, no previous work has used SNNs for skin cancer classification. Two deep CNNs, IRN and RN, were used as embedding layers, and were trained and tested on HAM10000, ISIC2019 and PAD-UFES-20. A shallower CNN was also used to compare depth versus classification performance. 

The classification results of each model have provided several insights. IRN performs best in terms of top-1 accuracy on the HAM10000 and ISIC2019 datasets, with results of 85.16\% and 78.39\%, respectively. These scores imply that the image uniformity in HAM10000 has improved performance. The far shallower CNN gives top-3 accuracy results similar to the deeper CNNs on the dermoscopic datasets, but performs significantly worse with clinical images in PAD-UFES-20. RN outperforms IRN on this dataset in both top-1 and top-3 accuracy, with scores of 74.33\% and 85.96\%, respectively. These results suggest that RN is more suitable for deployment on a mobile platform where the network is required to classify clinical images. When identifying new classes, the distance threshold outperforms probability in the majority of tests. However, the embedding space could only identify a maximum of 15.32\% of the new class skin lesion images without significantly impeding classification ability. Another test, using faces from LFW as the new class, identifies 62.45\% of new class images. This highlights the potential for this framework.

To build upon this work, alternative CNNs could be trialled as the embedding layer in the SNN. In particular, Inception-V4 has been effective when classifying skin disease datasets in previous work \cite{liuDeepLearningSystem2020, emaraModifiedInceptionv4Imbalanced2019}. Partial fine-tuning could also be evaluated. This may distinguish new classes more clearly using their low-level features, thereby making them more distinct in hyperspace. As a longer-term project, patient metadata could be used as an additional input to the SNN, as employed by Liu \etal \cite{liuDeepLearningSystem2020} and Krohling \etal \cite{krohlingSmartphoneBasedApplication2021}. 
The effect of sparsity in the embedding space, mentioned in our results and in previous work \cite{wangPlantLeavesClassification2019}, could be more thoroughly evaluated by thinning to varying degrees. Finally, a different ranking loss function, such as circle loss \cite{sunCircleLossUnified2020}, could be explored. 

\bibliographystyle{splncs04}
\bibliography{My-Library}

\end{document}